\documentclass[conference]{IEEEtran}
\IEEEoverridecommandlockouts
\usepackage{cite}
\usepackage{amsmath,amssymb,amsfonts}
\usepackage{algorithmic}
\usepackage{graphicx}
\usepackage{textcomp}
\usepackage{microtype}
\usepackage{graphicx}
\usepackage{booktabs} %
\usepackage{siunitx}
\usepackage{caption}
\usepackage{url}
\usepackage{hyperref}
\usepackage{mathrsfs}
\usepackage{todonotes}

\usepackage{subcaption}
\usepackage{multirow}
\usepackage{xcolor}

\usepackage{amsmath}
\usepackage{amsthm}
\usepackage{amssymb}

\usepackage{overpic}
\usepackage{bm}

\usepackage[switch]{lineno}

\usepackage{fancyhdr}
\renewcommand{\headrulewidth}{0pt}
\usepackage{xcolor}
\def\BibTeX{{\rm B\kern-.05em{\sc i\kern-.025em b}\kern-.08em
    T\kern-.1667em\lower.7ex\hbox{E}\kern-.125emX}}
\begin{document}

\title{Koopman Invertible Autoencoder: Leveraging Forward and Backward Dynamics for \\Temporal Modeling}

\author{\IEEEauthorblockN{Kshitij Tayal}
\IEEEauthorblockA{\textit{University of Minnesota}\\
tayal@umn.edu}
\and
\IEEEauthorblockN{Arvind Renganathan}
\IEEEauthorblockA{\textit{University of Minnesota}\\
renga016@umn.edu}
\and
\IEEEauthorblockN{Rahul Ghosh}
\IEEEauthorblockA{\textit{University of Minnesota}\\
ghosh128@umn.edu}
\and
\IEEEauthorblockN{Xiaowei Jia}
\IEEEauthorblockA{\textit{University of Pittsburgh}\\
xiaowei@pitt.edu}
\and
\IEEEauthorblockN{Vipin Kumar}
\IEEEauthorblockA{\textit{University of Minnesota}\\
kumar001@umn.edu}
}

\maketitle

\begin{abstract}

Accurate long-term predictions are the foundations for many machine learning applications and decision-making processes. However, building accurate long-term prediction models remains challenging due to the limitations of existing temporal models like recurrent neural networks (RNNs), as they capture only the statistical connections in the training data and may fail to learn the underlying dynamics of the target system. To tackle this challenge, we propose a novel machine learning model based on Koopman operator theory, which we call Koopman Invertible Autoencoders (KIA), that captures the inherent characteristic of the system by modeling both forward and backward dynamics in the infinite-dimensional Hilbert space. This enables us to efficiently learn low-dimensional representations, resulting in more accurate predictions of long-term system behavior. Moreover, our method's invertibility design guarantees reversibility and consistency in both forward and inverse operations. We illustrate the utility of KIA on pendulum and climate datasets, demonstrating 300\% improvements in long-term prediction capability for pendulum while maintaining robustness against noise. Additionally, our method excels in long-term climate prediction, further validating our method's effectiveness.

\end{abstract}

\section{Introduction}
Temporal data, prevalent in many applications such as climate, finance, and biomedicine, %
present a challenging problem for accurate long-term prediction and forecasting. %
Recurrent Neural Networks (RNNs) have gained significant attention for their ability to model sequential data by maintaining an internal time-evolving state.  %
However, %
a primary concern in training and deploying RNNs  is in their degraded performance %
over extended time horizons, which stems from the problem of exploding and vanishing gradients \cite{pascanu2013difficulty}. This gradient instability can result in slow convergence or even hinder learning completely, making it less suitable for capturing long-term dependencies in the data. To address this issue, researchers have proposed various methods, such as constraining the weight matrix to belong to the orthogonal group \cite{lezcano2019cheap}, using unitary hidden-to-hidden matrices \cite{kerg2019non}, temporal convolutional networks \cite{bai2018empirical}, and %
other solutions \cite{lai2018modeling}. Despite these efforts, achieving long-term memory is still an ongoing challenge and remains an active area of research.

Additionally, applying existing temporal models directly for scientific problems presents multiple obstacles: Firstly, accurate depiction of spatial and temporal processes within physical systems necessitates a substantial amount of training data \cite{siami2019performance}, which is often scarce in real-world situations. Secondly, existing empirical models remain limited in generalizing to scenarios that look different from training data. This is because they 
only establish statistical connections \cite{mello2018machine} between input and the targeted system variables %
but do not consider the inherent characteristics of the processes involved in the target system. %
Lastly, the relationships learned by these models are only valid for the specific distribution of forcing variables present in the training data, limiting their ability to generalize to scenarios not covered in the training set. In a study by Read et al.~\cite{read2019process}, it was demonstrated that an RNN model trained solely on data from a water body under current climatic conditions struggled to accurately predict outcomes in different climate scenarios, highlighting the limited generalizability of existing models in such cases.

In recent times, Koopman-based models have gained attention as a promising alternative approach for modeling temporal data \cite{brunton2022data}.
These models are based on the Koopman operator (\cite{koopman1931hamiltonian}, also see related work), transforming the original nonlinear system into an infinite-dimensional linear space. Koopman operator has three distinctive properties, which make it an ideal choice for temporal modeling (i) \emph{Linearity}~\cite{lan2013linearization}: The Koopman operator turns the original nonlinear system into an infinite-dimensional linear system, which simplifies the process of capturing the inherent patterns and trends in the data, which is crucial for effective temporal modeling. (ii) \emph{Global analysis} \cite{mezic2013analysis}: Unlike other linearization techniques (e.g., linearization around fixed points or periodic orbits), the Koopman operator provides a global perspective, capturing the overall behavior of the system rather than just local dynamic that can enhance the generalization capabilities of the model. (iii) \emph{Invariant properties}~\cite{takeishi2017learning}: The eigenfunctions and eigenvalues of the Koopman operator can reveal intrinsic properties of the system that remain unchanged under the system dynamics, which helps uncover hidden structures and patterns and makes it more robust to noise. These three distinctive properties of the Koopman operator make it a powerful tool for modeling temporal data.

However, using the Koopman operator for practical computations can be challenging because it is an infinite-dimensional operator~\cite{mauroy2020koopman}. Recently, researchers have developed techniques to approximate the Koopman operator using finite-dimensional representations extracted by autoencoder-based models, e.g.,  %
the Koopman Autoencoder (KAE) \cite{lusch2018deep}. %
These models effectively reduce the complexity by creating a low-dimensional representation space in which 
the Koopman operator can be suitably approximated with a linear layer that accurately captures the underlying dynamics of the system. However, because of model architecture, the information gleaned from Koopman-based models often is primarily based on forward dynamics, which overlooks the scope to acquire knowledge from backward dynamics.  
The fundamental goal of the forward run is to move from the present state to the subsequent state. Conversely, a backward run aims to go from the current state to the one that preceded it. Modeling backward dynamics ensures linearity in the low-dimensional space and also regularizes the forward run to be consistent.

One straightforward strategy is to have two separate linear layers \cite{azencot2019consistent}, each dedicated to independently modeling the forward and backward runs. However, such an approach may have a limited capacity to accurately capture the intrinsic dynamics of the process due to its need for knowledge sharing between the forward and backward states. 
This paper aims to build a long-term predictive model by leveraging  %
the Koopman analysis to capture both forward and backward dynamics in a unified model.   %
In particular, we model the forward and backward dynamics in the low-dimensional space using an invertible neural networks model ~\cite{kingma2018glow}, which can establish explicit invertible mappings between the input and output spaces. 
As a result of this integrated approach, a single layer can be trained to learn both forward and backward processes. This unified model can leverage common knowledge between the two directions and enhances the ability to capture the process dynamics fully. %
To the best of our knowledge, this paper is the first to present an invertibility approach for learning the Koopman operator. While analogous concepts have been utilized in the field of inverse modeling, i.e., using observable data to infer hidden characteristics, and vice versa \cite{tayal2022invertibility},  our research lays the foundation for predictive models that incorporate the essential underlying dynamics and go beyond mere trend prediction, which is of great significance in temporal modeling.

Our main contributions are as follows.
\begin{itemize}
	\item In this work, we present Koopman Invertible Autoencoders (KIA), a novel approach that harnesses both forward and backward dynamics for learning low-dimensional representations of temporal data and illustrates its utility on the pendulum and climate datasets. 
	
	\item We accurately extracted the pendulum system's dynamics, handling both clean and noisy scenarios, and achieved a remarkable  300\% improvement in long-term prediction accuracy.
	
	\item We demonstrated the capability of our model to comprehend the intricate dynamics of the climate dataset, which enables generalization across diverse weather scenarios and makes accurate long-term predictions.
\end{itemize}
\section{Related Work}

RNNs \cite{yu2019review} and their variants have proved indispensable in dealing with sequential data, making strides in numerous applications such as language modeling \cite{kadar2017representation}, speech recognition \cite{karita2019comparative}, and time-series  prediction\cite{lim2021time}. However, despite the versatility of RNNs, they are plagued by issues such as vanishing and exploding gradients \cite{bengio1994learning}, hindering their ability to model long-term dependencies. Variants of RNNs such as Long Short-Term Memory (LSTM) \cite{hochreiter1997long}, Gated Recurrent Unit (GRU) \cite{dey2017gate}, and Quasi-Recurrent Neural Networks (QRNN) \cite{bradbury2016quasi} have been proposed to overcome this difficulty and have achieved remarkable results. These networks address the limitations of RNN by utilizing various gating mechanisms allowing them to control the information flow, retain long-term dependencies, and mitigate vanishing/exploding gradient problems.

However, RNN-based models remain limited for long-term prediction as they rely on complex non-linear temporal structures (e.g., LSTM unit)  and can easily accumulate errors over time. 
A recent trend in machine learning is to explore %
transferring  physical theories from existing physics-based models to enhance the capabilities and generalization of RNNs~\cite{willard2022integrating} 
. It combines the best of both worlds: the structure and explanatory power of physics-based models and the learning and predictive capabilities of RNNs. For instance, a physics-guided recurrent neural network model (PGRNN)  \cite{jia2019physics} %
integrated the heat transfer process with the RNN model to effectively captured the long-term dependencies in the data, a task at which traditional RNNs failed.
Other works also leveraged physics to enhance the long-period prediction of turbulent flows~\cite{wang2020towards,bao2022physics}.  However, these methods rely on thorough knowledge of the system's physics, which limits their applicability. Furthermore, current model architectures, both generic and physics-based,  predominantly rely on using forward dynamics for training, neglecting the potential of backward dynamics. This lack of integration due to network architecture presents an overlooked opportunity for enhancing these systems. In contrast, our approach builds upon the dynamical systems theory to assimilate both forward and backward dynamics through which it learns the true underlying processes. This significantly improves long-term prediction capabilities and offers an alternative to prevailing methodologies.

The theory of Koopman operators \cite{koopman1931hamiltonian}, established in 1931, has recently emerged as a groundbreaking framework for systematic linearization of complex dynamical system within an infinite-dimensional Hilbert space. Dynamic Mode Decomposition (DMD)\cite{schmid2010dynamic} and Extended Dynamic Mode Decomposition (EDMD) \cite{williams2015data} are two popular approaches for approximating this operator; however, they face issues of computational feasibility\cite{brunton2022data}. To tackle these challenges, researchers have begun to utilize data-driven approaches through deep learning to determine the Koopman operator from observed data \cite{takeishi2017learning,lusch2018deep}. Most of these strategies employ autoencoders \cite{wang2023koopman} to transition from nonlinear to linear Koopman subspaces in order to identify the Koopman operator. For instance, the VAMP architecture \cite{mardt2018vampnets} utilizes a time-lagged autoencoder and a custom variational score to find Koopman coordinates on a protein folding example. Similarly, researchers \cite{berman2023multifactor} used koopman operator for sequence disentanglement by assuming that the underlying complex dynamics can be represented linearly in latent space. The focus of our work is on long-term temporal modeling whereby we want to model both forward and backward dynamics, which is significantly more manageable in a linearized space as backward dynamics are the inverse of forward space. A method most relevant to our approach is C-KAE \cite{azencot2020forecasting}, where the authors developed two separate networks for learning forward and backward dynamics, which are then trained together using consistency loss. Nevertheless, training two separate networks overlooks the connection between forward and backward dynamics and consequently loses the opportunity to exploit shared knowledge. Additionally, optimizing two neural network to be inverse of each other is computationally difficult and unstable due to their stochastic nature. Recently, a new category of neural networks, called Invertible Neural Networks (INNs), was introduced \cite{dinh2014nice,kingma2018glow}, based on the principles of normalizing flow. INNs are bijective functions that, by design, can simultaneously be trained on both forward and backward dynamics and exploit shared knowledge, making them an ideal choice for long-term temporal modeling.

\begin{figure*}[!t]
	\centering
	\begin{subfigure}[t]{0.50\textwidth}
		\centering
		\begin{overpic}[width=1\textwidth]{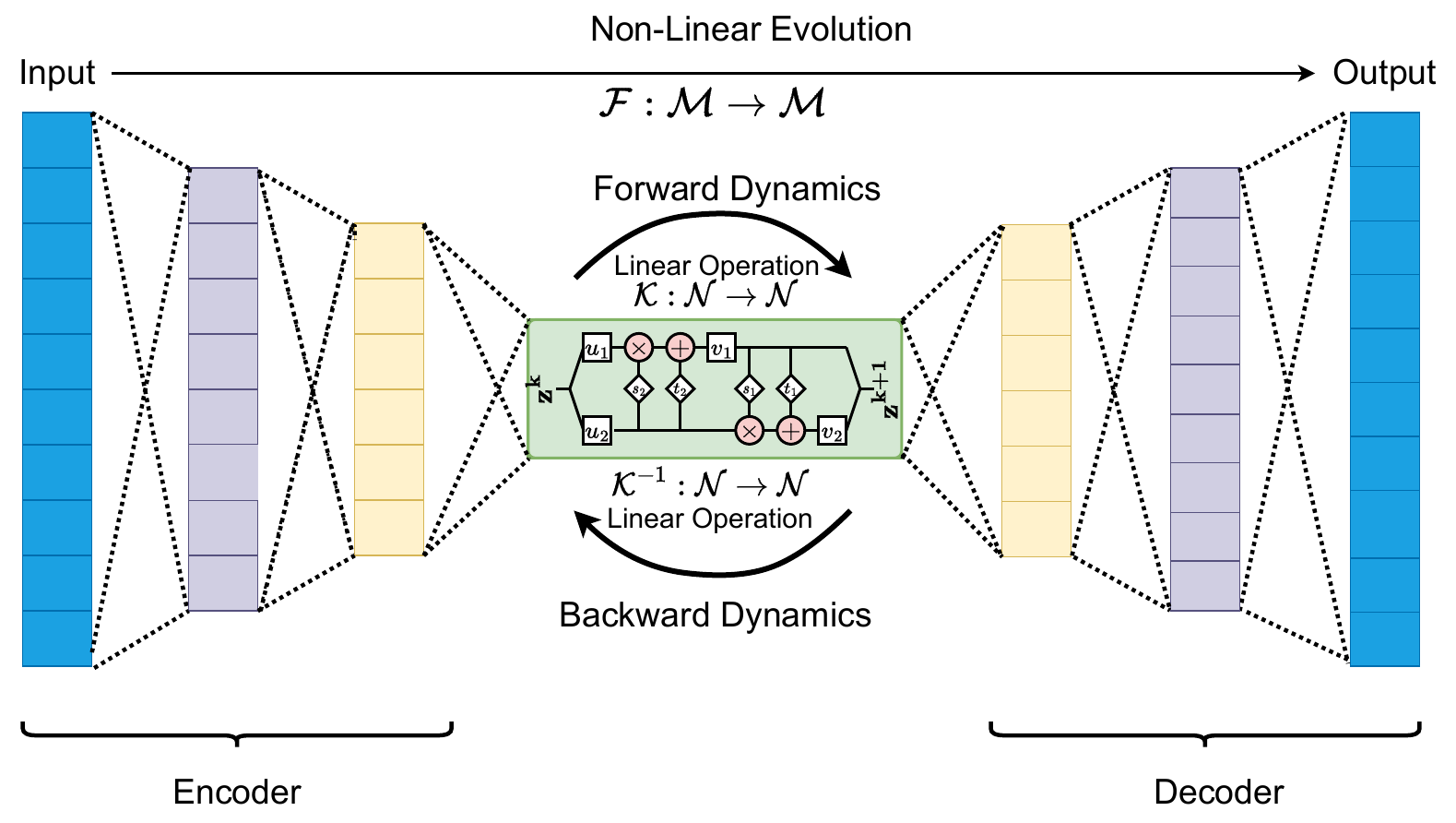} 
		\end{overpic}		
		
		\caption{\small KIA Architecture}\label{fig:pendulum_1_noise_a}
	\end{subfigure}%
	~
	\begin{subfigure}[t]{0.46\textwidth}
		\centering
			
		\begin{overpic}[width=1\textwidth]{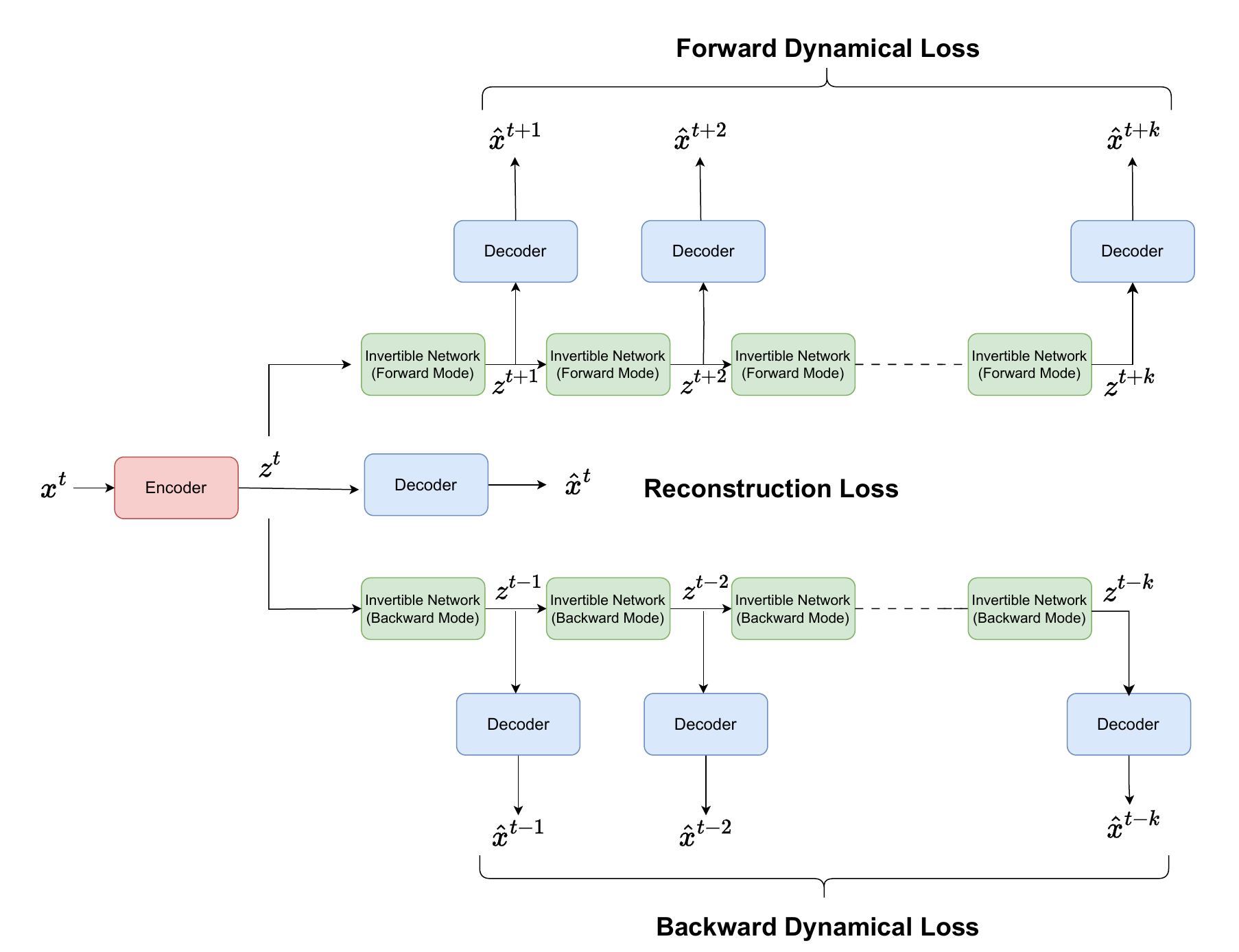} 
		\end{overpic}
	\captionsetup{justification=justified}
         \caption{\small Illustration of Different Loss Functions}\label{fig:pendulum_1_noise_c}
           
	\end{subfigure}%
	
	\caption{\small The left figure depicts our network architecture (KIA). The right figure showcases the various losses employed where observations $\bm{x}_t$ are inputted and transformed into a latent representation $\bm{z}_t$ through an encoder and propagated forward and backward using INNs}
	\label{fig:koopman_inn_architecture}
\end{figure*}

\section{Problem Formulation \& Preliminaries}
\label{subsec:pbm_def}

\subsection{Problem Formulation}
Temporal data, denoted as $\{\bm{x}_t\}_{t=1}^{T}$, can be interpreted as a series of observations from a dynamical system. This could represent anything that varies over time, for example, weather predictions. stock price etc. Consider the following discrete form of the system  dynamics
\begin{align} \label{eq:disc_dyn}
	\bm{x}_{t+1} = \mathbf{\mathscr{F}}(\bm{x}_t) + \bm{r}_t \ , \quad \bm{x} \in \mathcal{M} \subset \mathbb{R}^m \ ,
\end{align}

 where $\bm{x}_{t+1}$ represents the state of the system at the next time step, given its current state $\bm{x}_t$ and $\bm{r}_t \in \mathcal{M}$ represents deviation from the true dynamics due to e.g., measurement errors or missing values. Here, the function $\mathbf{\mathscr{F}}(\bm{x}_t)$ is an (possibly non-linear) update rule which describes how the state of the system evolves from one time step to the next. Additionally, $\mathcal{M}$ represents a finite-dimensional manifold, embedded in a higher-dimensional Euclidean space, denoted by $\mathbb{R}^m$. In this work, we focus on multi-step forecasting task of predicting the future observations given the current observation. Formally, we seek to learn function map $\mathbf{\mathscr{F}}$ such that 
\begin{align} \label{eq:multistep_pushfwd}
	\bm{x}_{t+l} = \bm{x}_t \circ \mathbf{\mathscr{F}}^l \ , \quad l = 1,2,... \ ,
\end{align}
where $\bm{x}_{t+l}$ signifies the state of the system at a future time step and $\circ$ denotes function composition. $\mathbf{\mathscr{F}}^l$ indicates that we're applying the system dynamics repeatedly, $l$ times in total, to the current state $\bm{x}_t$. The above model assumes that future states $\bm{x}_{t+l}$ depend only on the current observation $\bm{x}_t$ and not on the information from a sequence of previous observations (between $t$ and $t+l$).

\subsection{Linear Invertible Neural Network (INNs) } INNs are bijective functions with a forward mapping $\mathcal{K} : \mathbb{R}^{n} \rightarrow \mathbb{R}^{n}$ and an inverse mapping $\mathcal{K}^{-1}: \mathbb{R}^{n} \rightarrow \mathbb{R}^{n}$. These mappings can be computed in closed-form manner , building upon the foundational works by Dinh et al., \cite{dinh2014nice,kingma2018glow}. To construct a linearly invertible neural network, we utilize the framework of the real non-volume preserving architecture proposed by \cite{dinh2016density}. The basic unit of this network comprises a reversible bijective network, incorporating two interconnected linear coupling layers. During the forward processing stage, as described by equation (\ref{eqn:inn_forward}), the input vector, denoted as $\boldsymbol{u}$, is partitioned into $\boldsymbol{u}_1$ and $\boldsymbol{u}_2$. These halves are then linearly transformed using translation functions $t_i(\cdot)$, where $i \in {1,2}$, and they are modeled by linear neural networks trained in the forward pass. The output are the concatenation $[\boldsymbol{v}_1, \boldsymbol{v}_2]$, which are computed as 
\begin{align}
\label{eqn:inn_forward}
    \begin{split}
        \boldsymbol{v}_1 &= \boldsymbol{u}_1  + t_2(\boldsymbol{u}_2)\\
        \boldsymbol{v}_2 &= \boldsymbol{u}_2 + t_1(\boldsymbol{v}_1)
    \end{split}
\end{align} 

Given the output $\boldsymbol{v} = [\boldsymbol{v}_1, \boldsymbol{v}_2]$, the above expressions are easily invertible as follows:
\begin{align}
    \label{eqn:inn_reverse}
    \begin{split}
        \boldsymbol{u}_2 = \boldsymbol{v}_2 - t_1(\boldsymbol{v}_1)  \\
        \boldsymbol{u}_1 = \boldsymbol{v}_1 - t_2(\boldsymbol{u}_2)
    \end{split}
\end{align} 

Importantly, the mappings $t_i$ can be arbitrarily complex functions. %
In our implementation, we implement bijectivity by a fully connected linear layers with linear activations and use deep INN  which is composed of a sequence of these reversible blocks. 
One advantage of INNs is that can we can train them on the well-understood forward process $\mathcal{K}$ and get the inverse $\mathcal{K}^{-1}$ for free by running them backward. This provides a unique opportunity to develop a unified model that integrates both forward and inverse dynamical processes seamlessly.

\subsection{Koopman Operator Theory}
The Koopman theory proposes a framework that allows us to study complex systems in a linear space. According to this theory, every nonlinear dynamical system can be transformed into a form where the evolution of observations can be described by an infinite-dimensional Koopman operator. This operator operates on the space of all possible measurement functions, which can be fully represented through a linear map. Formally, we define the Koopman operator as $\mathcal{K}_{\mathbf{\mathscr{F}}}^{\infty} : \mathcal{G}(\mathcal{M}) \rightarrow \mathcal{G}(\mathcal{M})$, where $\mathcal{G}(\mathcal{M})$ is the space of real-valued measurement functions, represented by $g: \mathcal{M} \rightarrow \mathbb{R}$. The Koopman operator maps between function spaces and transforms the observations of the state to the next time step. Mathematically, the action of the Koopman operator on a measurement function $g$ at time $t$ is given by:
\begin{align}
\label{eq:koopman}
\mathcal{K}_{\mathbf{\mathscr{F}}}^{\infty} g(\bm{x}_t) = g(\mathcal{K}_{\mathbf{\mathscr{F}}}^{\infty}(\bm{x}_t)),
\end{align}
where $\mathbf{\mathscr{F}}(\bm{x}_t)$ represents the transformed state of the system to the next time step. Also, the Koopman operator is a linear operator. This means that for any real numbers $\alpha$ and $\beta$, the following linearity property holds:
\begin{align*}
\mathcal{K}_{\mathbf{\mathscr{F}}}^{\infty} (\alpha g_1 + \beta g_2) &= (\alpha g_1 + \beta g_2) \circ \mathcal{K}_{\mathbf{\mathscr{F}}}^{\infty} \\
&= \alpha g_1 \circ \mathcal{K}_{\mathbf{\mathscr{F}}}^{\infty} + \beta g_2 \circ \mathcal{K}_{\mathbf{\mathscr{F}}}^{\infty} \\
&= \alpha \mathcal{K}_{\mathbf{\mathscr{F}}}^{\infty} g_1 + \beta \mathcal{K}_{\mathbf{\mathscr{F}}}^{\infty} g_2 \
\end{align*}
This property ensures that the Koopman operator preserves the linear structure of the measurement functions' space.

\section{Koopman Invertible Autoencoder (KIA)}
For forecasting future observations, an intuitive route is to design a neural network that effectively learns an approximation of the mapping denoted by $\mathbf{\mathscr{F}}$. However, a model developed using this approach ignores the underlying dynamics and can be hard to analyze. To address these challenges, we propose to develop the Koopman Invertible Autoencoder (KIA), which  is built upon the Koopman theory and INNs. 
In the following, we first provide an overview of the Koopman Autoencoder and how it incorporates the concepts from the Koopman theory. Subsequently, we will describe the integration of INNs, focusing on how they facilitate bidirectional modeling and contribute to the robustness of our proposed model.

\subsection{Koopman Autoencoder (KAE)}
Due to the infinite-dimensional nature of $\mathcal{K}_{\mathbf{\mathscr{F}}}^{\infty}$ in Eq. \eqref{eq:koopman}, %
the practical use of the Koopman theory requires creating a finite-dimensional approximation
that is capable of encapsulating the majority of the dynamics at play. We denote  finite-dimensional Koopman operator by $\mathcal{K}_{\mathbf{\mathscr{F}}}$.  The primary objective of the KAE model lies in the discovery of the matrices $\mathcal{K}_{\mathbf{\mathscr{F}}}$ and the nonlinear transformation of observation  to the Koopman subspace that allows for an accurate recovery of the underlying dynamics. The Koopman autoencoder consists of an encoder $\gamma_e$, which maps observations to the Koopman subspace, and a decoder $\gamma_d$, which maps the Koopman subspace back to the observation space. Both encoder and decoder are represented by neural networks. %
The training of the networks is conducted by minimizing the discrepancy between the input (${\bm{x}_t}$) and its corresponding output ($\tilde{\bm{x}_t} \&= \gamma_d \circ \gamma_e(\bm{x}_t)$) .  This guarantees that the encoder and decoder are solely responsible for encoding and decoding processes and do not learn any dynamics. This  %
can be formally represented as $\gamma_d \circ \gamma_e \approx \mathbb{I}$, where $\mathbb{I}$ stands for the identity function. The loss function to train this network is given by:
\begin{align}
    \label{eq:loss_id}
    \mathcal{L}_{Recon} &=  \frac{1}{n}\sum_{t=1}^n \| \gamma_d \circ \gamma_e(\bm{x}_t) -  \bm{x}_t \|_2^2 \ ,
\end{align}
which measures the mean squared error between the reconstructed and original data points over a dataset of size $n$.

 \begin{figure*}[t]
    \begin{subfigure}{\linewidth}
        \centering
        \includegraphics[width=\linewidth]{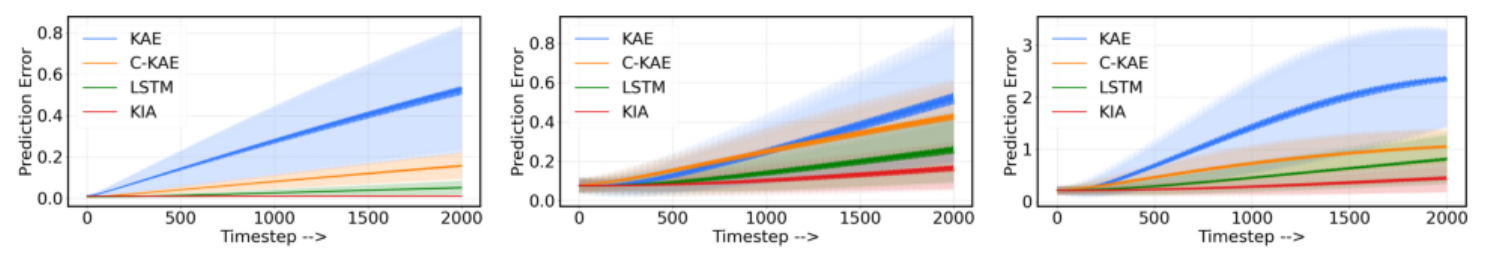}
        \label{fig:base}
    \end{subfigure}    
    \begin{subfigure}{\linewidth}
        \centering
        \includegraphics[width=\linewidth]{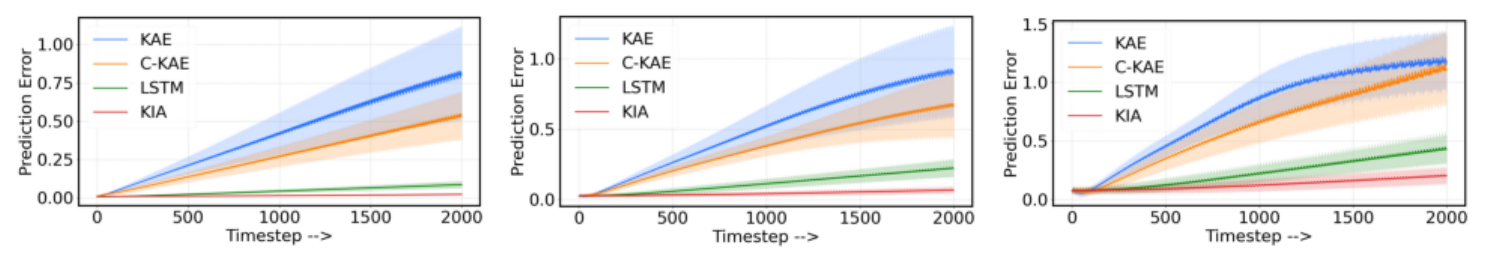}
        \label{fig:adapt}    
    \end{subfigure}
    \caption{\small Prediction errors over a time horizon of 2000 steps for clean and noisy pendulum observations with initial conditions $\theta = 0.8$ (top row) and  $\theta = 2.4$ (bottom row). The first column shows the clean results, the second column shows results with small noise, and the third column shows results with large noise. (Best seen in color)}
    \label{fig:pendulum_prediction}
\end{figure*}

\paragraph*{Modeling Forward Dynamics}
In general, $\mathcal{K}_{\mathbf{\mathscr{F}}}$ prescribes a rule to move forward in time i.e. $\hat{\bm{x}}_{t+1} \&= \gamma_d \circ \mathcal{K}_{\mathbf{\mathscr{F}}} \circ \gamma_e (\bm{x}_t)$. We can leverage this relationship for multi-step forecasting as : $\gamma_d \circ \mathcal{K}_{\mathbf{\mathscr{F}}}^{l} \circ \gamma_e(\bm{x}_t) \approx \bm{x}_t \circ \mathbf{\mathscr{F}}^l$, i.e., we iteratively obtain forward estimates. In order to accurately represent these forward dynamics, we apply a linear invertible neural network to serve as the approximate Koopman operator, denoted as $\mathcal{K}_{\mathbf{\mathscr{F}}}$. Our encoding network, denoted as $\gamma_e$, processes observations $\bm{x}_t$ to acquire a latent representation $\bm{z}_t$, in which the dynamics become linear through the encoder structure. During the forward evolution step (Eq \ref{eqn:inn_forward}), the input vector, referred to as $\bm{z}_t$, is divided into two equal parts, $\bm{z}_{t_1}$ and $\bm{z}_{t_2}$, which are then propagated forward as follow:
\begin{align}
\label{eq:beautified}
    \begin{split}
        \mathbf{z}_{t+1,1} &= \mathbf{z}_{t,1} + t_2(\mathbf{z}_{t,2}) \\
        \mathbf{z}_{t+1,2} &= \mathbf{z}_{t,2} + t_1(\mathbf{z}_{t+1,1})
    \end{split}
\end{align} where the concatenation of output $[ \mathbf{z}_{t+1,1} ,\mathbf{z}_{t+1,2}]$ from INNs represent the forward evolution in Koopman subspace.  In our tests, we noticed that our models predict as well as generalize better in %
multi-step forecasting, compared with the traditional auto-regressive approach that %
computes one step forward at a time. Given a choice of $k$ forward prediction steps, we define the following forward dynamical loss term:
\begin{align}
\label{eq:loss_pred_fwd}
\mathcal{L}_{\mathrm{fwd}} &= \frac{1}{ k * n} \sum_{l=1}^{k} \sum_{t=1}^n \|  \gamma_d \circ \mathcal{K}_{\mathbf{\mathscr{F}}}^{l} \circ \gamma_e(\bm{x}_t)   - \bm{x}_{t+l} \|_2^2 \ ,
\end{align}

\subsection{Bidirectional Modeling}
The majority of Koopman-based networks currently available \cite{takeishi2017learning,lusch2018deep} primarily focus on modeling forward dynamics. 
A benefit of preserving linearity in the latent space is that the evolution matrix $\mathcal{K}_{\mathbf{\mathscr{F}}}$ can also be exploited for backward prediction via its inverse: $\mathcal{K}_{\mathbf{\mathscr{F}}}^{-1}$, i.e., $\bar{\bm{x}}_{t-1} = \gamma_d \circ \mathcal{K}_{\mathbf{\mathscr{F}}}^{-1} \circ \gamma_e(\bm{x}_t)$. KAE architectures that employ the liner layer \cite{takeishi2017learning,lusch2018deep} to learn forward dynamics cannot learn the backward dynamics due to the high computational cost of matrix inversion, which is typically $O(n^3)$. Another approach to implementing backward dynamics is through independent linear layers~\cite{azencot2020forecasting}, but this may limit the capacity to capture the intrinsic dynamics due to the lack of knowledge sharing between the forward and backward states. Earlier methods have also examined the integration of backward dynamics into their non-linear model, such as in bi-directional RNN~\cite{schuster1997bidirectional}. However, the inherent nonlinearities of a typical neural network make it difficult to constrain the forward and backward models. 

Our tests found that models trained for forward prediction typically produce poor, backward predictions. In contrast, considering both forward and backward dynamics can contribute to more effective training of KAE. Specifically, incorporating backward dynamics ensures linearity in the low-dimensional space while regularizing the consistency with the forward run. To address this limitation, we propose modeling the forward and backward dynamics in the low-dimensional space using INNs which are invertible by design and very efficient. In contrast to existing methods that use separate structures and parameters for backward modeling,
our model allows for the direct back prediction through the INN structure (Eq.~\ref{eqn:inn_reverse}). Specifically, the  latent representation $\bm{z}_t$ ($\gamma_e \circ \bm{x}_t$) is divided into two parts, $\bm{z}_{t_1}$ and $\bm{z}_{t_2}$, which are then propagated backward as follow:
\begin{align}
    \begin{split}
        \mathbf{z}_{t-1,2} = \mathbf{z}_{t,2} - t_1(\mathbf{z}_{t,1})  \\
        \mathbf{z}_{t-1,1} = \mathbf{z}_{t,1} - t_2(\mathbf{z}_{t-1,2})
    \end{split}
\end{align} where the concatenation of output $[ \mathbf{z}_{t-1,1} ,\mathbf{z}_{t-1,2}]$ from INNs represent the backward evolution in Koopman subspace. As with forward dynamics, we noticed that our models predict as well as generalize better in multi-step backcasting. Given a choice of $k$ backward prediction steps, we define the following backward dynamical loss term:
\begin{align}
\label{eq:loss_pred_bwd}
\mathcal{L}_{\mathrm{bwd}} &= \frac{1}{ k * n} \sum_{l=1}^{k} \sum_{t=1}^n \|  \gamma_d \circ \mathcal{K}_{\mathbf{\mathscr{F}}}^{-l} \circ \gamma_e(\bm{x}_t)   - \bm{x}_{t-l} \|_2^2 \ ,
\
\end{align}

We derive our KIA framework for analyzing temporal data by integrating all the above components. Our model undergoes training by minimizing a loss function whose minimizers guarantee that we achieve an optimized autoencoder and get accurate predictions over time by effectively capturing both forward and backward dynamics. We define the complete training loss as  %
\begin{align} \label{eq:ours_loss}
\mathcal{L} = \lambda_{\mathrm{Recon}} \mathcal{L}_{\mathrm{Recon}} + \lambda_{\mathrm{fwd}} \mathcal{L}_{\mathrm{fwd}} + \lambda_{\mathrm{bwd}} \mathcal{L}_{\mathrm{bwd}}\ ,
\end{align}
where $\lambda_\mathrm{Recon}, \lambda_\mathrm{fwd}, \lambda_\mathrm{bwd}$, are parameters that balance between reconstruction, forward and backward prediction.  
Figure \ref{fig:koopman_inn_architecture} (left) depicts our network design with an encoder, decoder, and an INN-based Koopman module for forward and backward dynamics computation. Figure \ref{fig:koopman_inn_architecture} (right) illustrates our loss landscape.

\section{Experiments}
\label{sec:experiments}
To evaluate our proposed approach for long term temporal modeling, we perform a comprehensive study  on two datasets, i.e., the pendulum dataset and the sea surface temperature dataset, and compare with state of the art Koopman-based approaches as well as baseline sequential models.

\subsection{Nonlinear Pendulum}
\label{subsec:pendulum}

\textbf{Dataset Description:} 
We evaluate our models on a non-linear pendulum, whose behavior is described by a second-order ordinary differential equation (ODE), where the angular displacement from equilibrium, denoted by $\theta$, follows the equation $\frac{d^2\theta}{dt^2} = -\frac{g}{l}\sin(\theta)$. We use standard parameters for the length, $l$ (1), and gravity's acceleration, $g$ ($9.8$ $m/s^2$). The pendulum's behavior is different based on the initial angle. A small initial angle results in simple harmonic motion, while a larger initial angle introduces non-linearity and complexity. With this in mind, we conducted experiments using two specific initial angles for oscillations, $\theta = 0.8$ and $\theta = 2.4$, over a period ranging from $0$ to $400$. We also employ a random orthogonal transformation in $\mathbb{R}^{64\times 2}$ to convert our input data points into a high-dimensional space. This transformation results in observations that reside in $\mathbb{R}^{64}$, enhancing the representation of the original data. We collected 4,000 evenly spaced points in $\mathbb{R}^2$ over the time interval $t = [0,400]$. This dataset was divided into training (400 points), validation (1,500 points), and testing (2,100 points) sets. The model was trained on the training set, hyperparameter optimized using the validation set, and the final resulting model was evaluated on the test set.

\begin{table}[]
\centering
 \caption{\small Pendulum Results: KIA model shows significant error reduction, especially in the last 100 predictions, demonstrating effective capture of long-range temporal dependencies.}
 \resizebox{0.49\textwidth}{!}{
\begin{tabular}{ccccc}
\toprule
\textbf{Model} & \boldmath{$\theta$} & \multicolumn{3}{c}{\textbf{Prediction Error (Avg)}} \\
\cmidrule(lr){3-5}
& & \textbf{All Value} & \textbf{First 100}  & \textbf{Last 100} \\  
\midrule
KAE & 0.8 & 0.273 $\pm$ 0.151 & 0.020 $\pm$ 0.005 & 0.514 $\pm$ 0.014 \\
C-KAE & 0.8 & 0.080 $\pm$ 0.044 & 0.009 $\pm$ 0.001 & 0.154 $\pm$ 0.003 \\
RNN & 0.8 & 0.026 $\pm$ 0.013 & \textbf{0.008} $\pm$ 0.001 & 0.050 $\pm$ 0.001 \\
KIA & 0.8 & \textbf{0.010} $\pm$ 0.002 & 0.011 $\pm$ 0.001 & \textbf{0.011} $\pm$ 0.001 \\
\midrule
KAE & 2.4 & 0.416 $\pm$ 0.236 & 0.024 $\pm$ 0.010 &0.794 $\pm$ 0.015 \\
C-KAE & 2.4 & 0.270 $\pm$ 0.153 & 0.022 $\pm$ 0.006 & 0.524 $\pm$ 0.010 \\
RNN & 2.4 & 0.043 $\pm$ 0.023 & \textbf{0.008} $\pm$ 0.001 & 0.083 $\pm$ 0.001 \\
KIA & 2.4 & \textbf{0.014} $\pm$ 0.004 & 0.009 $\pm$ 0.001 & \textbf{0.022} $\pm$ 0.001 \\
\bottomrule 
\end{tabular}}
\label{tab:pendulum_results_1}
\end{table}

\begin{table}[h]
\caption{\small Range of parameter values tried for hyperparameter tuning, with the final selected value shown in \textbf{bold}.}
\resizebox{\linewidth}{!}{
\begin{tabular}{|l|l|}
\hline
\textbf{Hyperparameter}                     & \textbf{Value}             \\ \hline
Latent dimension of Encoder                 & 4, \textbf{8}, 16                 \\ \hline
Batch size                                  & 32, \textbf{64}, 128, 256                \\ \hline
Learning rate                               & 0.005, \textbf{0.001}, 0.0005, 0.01 \\ \hline
$\lambda_\mathrm{Recon}$, $\lambda_\mathrm{fwd}$                          & \textbf{1},2,3,4,5 \\ \hline
$\lambda_\mathrm{bwd}$  & 0.1,\textbf{0.5},1,2,5 \\ \hline
\end{tabular}
}
\label{tab:hyperparameter_tuning}
\end{table}

\textbf{Baselines:} \underline{\textit{Koopman Auto Encoder (KAE)} \cite{lusch2018deep}}: This network model approximates the Koopman operator by only looking at its forward dynamics. \underline{\textit{Consistent Koopman Auto Encoder (C-KAE)}\cite{azencot2020forecasting}}: This network model builds upon the original KAE and incorporates both forward and backward dynamics by using two separate Koopman operators. Further they enforce consistency between these operators by approximating one to be inverse of the other through consistency loss. \underline{\textit{Long Short-Term Memory (LSTM)}}: We further conducted a comparison against an LSTM, a variant of a recurrent neural network (RNN). This model is designed to learn a nonlinear function $\bm{x}_t$ $\rightarrow$ $\bm{x}_{t+1}$ directly. During the inference phase, this function takes $\bm{x}_t$ as input to predict $\bm{x}_{t+1}$, $\bm{x}_{t+2}$....$\bm{x}_{t+k}$ recursively. We use two LSTM layers, each with a  hidden dimension of 64.

All Koopman-based models, including our proposed method, utilize the same encoder-decoder architecture. Specifically, we employ a three-layer feed-forward network (128, 64, 8) with non-linear activations. While specialized architectures like convolutional neural networks with non-linear layers are another option, previous research has shown that these architectures do not offer any advantages for Koopman-based approaches \cite{rice2020analyzing}. All models underwent training for 500 epochs, utilizing Adam optimizers with identical learning rates. The training, validation, and test sets were consistent across all models. An early stopping criterion was also applied, terminating training if the validation loss did not decrease for 20 consecutive epochs. Moreover, the models were trained using the same random seeds to ensure consistency. In order to determine the optimal hyperparameters, we conducted an extensive grid search, exploring a range of parameter values specified in Table \ref{tab:hyperparameter_tuning}.

\begin{table}[]
\centering
\caption{\small Results for pendulum with Low noise.}
\resizebox{0.49\textwidth}{!}{
\begin{tabular}{ccccc}
\toprule
\textbf{Model} & \boldmath{$\theta$} & \multicolumn{3}{c}{\textbf{Prediction Error (Avg)}} \\
\cmidrule(lr){3-5}
& & \textbf{All Value} & \textbf{First 100}  & \textbf{Last 100} \\  
\midrule
KAE & 0.8 & 0.262 $\pm$ 0.140 & 0.078 $\pm$ 0.001 & 0.508 $\pm$ 0.020 \\
C-KAE & 0.8 & 0.250 $\pm$ 0.106 & 0.082 $\pm$ 0.003 & 0.421 $\pm$ 0.012 \\
RNN & 0.8 & 0.149 $\pm$ 0.058 & \textbf{0.076} $\pm$ 0.001 & 0.253 $\pm$ 0.012 \\
KIA & 0.8 & \textbf{0.109} $\pm$ 0.028 & 0.077 $\pm$ 0.001 & \textbf{0.162} $\pm$ 0.010 \\
\midrule
KAE & 2.4 & 0.500 $\pm$ 0.274 & 0.029 $\pm$ 0.007 & 0.901 $\pm$ 0.014 \\
C-KAE & 2.4 & 0.367 $\pm$ 0.195 & 0.032 $\pm$ 0.004 & 0.662 $\pm$ 0.010 \\
RNN & 2.4 & 0.113 $\pm$ 0.060 & \textbf{0.026} $\pm$ 0.001 & 0.217 $\pm$ 0.004 \\
KIA & 2.4 & \textbf{0.042} $\pm$ 0.012 & \textbf{0.026} $\pm$ 0.001 & \textbf{0.065} $\pm$ 0.001 \\
\bottomrule 
\end{tabular}}
\label{tab:pendulum_results_2}
\vspace{0.2cm}
\end{table}

\begin{table}[]
\centering
\caption{\small Results for pendulum with High noise.}
\resizebox{0.49\textwidth}{!}{
\begin{tabular}{ccccc}
\toprule
\textbf{Model} & \boldmath{$\theta$} & \multicolumn{3}{c}{\textbf{Prediction Error (Avg)}} \\
\cmidrule(lr){3-5}
& & \textbf{All Value} & \textbf{First 100}  & \textbf{Last 100} \\  
\midrule
KAE & 0.8 & 1.347 $\pm$ 0.715 & 0.211 $\pm$ 0.007 & 2.337 $\pm$ 0.033 \\
C-KAE & 0.8 & 0.687 $\pm$ 0.261 & 0.224 $\pm$ 0.010 & 1.042 $\pm$ 0.017 \\
RNN & 0.8 & 0.468 $\pm$ 0.188 & 0.214 $\pm$ 0.005 & 0.794 $\pm$ 0.019 \\
KIA & 0.8 & \textbf{0.298} $\pm$ 0.075 & \textbf{0.210} $\pm$ 0.005 & \textbf{0.440} $\pm$ 0.023 \\
\midrule
KAE & 2.4 & 0.757 $\pm$ 0.362 & \textbf{0.063} $\pm$ 0.008 & 1.173 $\pm$ 0.016 \\
C-KAE & 2.4 & 0.619 $\pm$ 0.322 & 0.066 $\pm$ 0.005 & 1.102 $\pm$ 0.021 \\
RNN & 2.4 & 0.227 $\pm$ 0.111 & 0.077 $\pm$ 0.004 & 0.421 $\pm$ 0.009 \\
KIA & 2.4 & \textbf{0.126} $\pm$ 0.039 & 0.076 $\pm$ 0.003 & \textbf{0.198} $\pm$ 0.004 \\
\bottomrule 
\end{tabular}}
\label{tab:pendulum_results_3}
\end{table}

\begin{table}[]
\centering
\caption{\small Varying amount of training data for pendulum with $\theta_0 = 2.4$. }
\resizebox{0.38\textwidth}{!}{
\begin{tabular}{ccc}
\toprule
\textbf{Model} & \textbf{Training Points} & \textbf{Prediction Error (Avg)} \\
\midrule
RNN & 200 & 0.129 $\pm$ 0.06\\
KIA & 200 & \textbf{0.052} $\pm$ 0.04\\
\midrule
RNN & 300 & 0.087 $\pm$ 0.06\\
KIA & 300 & \textbf{0.035} $\pm$ 0.03\\
\bottomrule 
\end{tabular}}
\label{tab:varying_training_points}
\end{table}

\textbf{Experimental results.} \
To evaluate prediction performance, we use scale invariant relative forecasting error at each timestep which is $\frac{\|\hat{\bm{x}}_t -\bm{x}_t \|_2}{\|\bm{x}_t\|_2}$.  In order to assess the long-term forecasting capabilities of our models, we perform forecasts over a horizon of 2,000 steps, using 30 different initial observations. The mean and standard deviation of prediction errors for these initial observations are presented in Table \ref{tab:pendulum_results_1}. Additionally, we provide insights into prediction capabilities across different time spans by reporting the average prediction error over all timesteps, the first 100 timesteps, and the last 100 timesteps. 

The average predictions for condition $\theta_0 = 0.8$ are consistently lower than for $\theta_0 = 2.4$, indicating that $\theta_0 = 2.4$ introduces more complexity and non-linearity. The RNN model performs better in the first 100 timesteps, suggesting its strength in capturing short-term temporal dynamics. Still, the KIA model shows a remarkable  300\% improvement in long-term prediction accuracy over RNN on average. We attribute this improvement to the KIA model's ability to capture long term trends by learning inherent characteristics via mapping forward and backward dynamics in a single network. C-KAE, which considers both forward and backward, outperforms the KAE, which emphasizes the importance of studying backward dynamics but lacks consistency as it uses separate layers.

\begin{figure}[!tbp]
    \begin{subfigure}{0.9\linewidth}
        \centering
        \includegraphics[width=\linewidth]{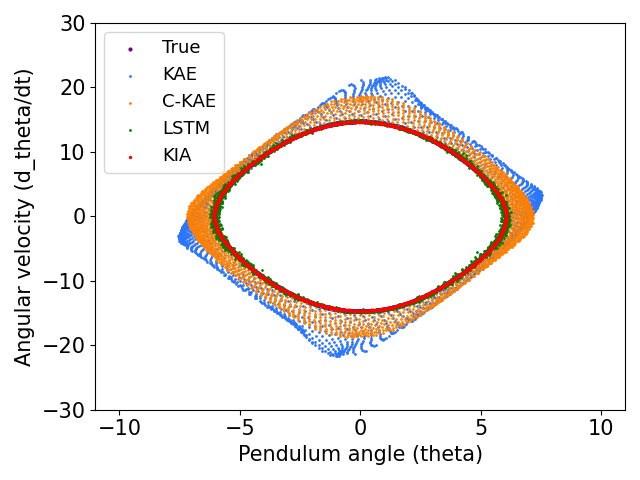}
        \label{fig:base}
    \end{subfigure}  
    \vspace{-0.5cm}
    \begin{subfigure}{0.9\linewidth}
        \centering
        \includegraphics[width=\linewidth]{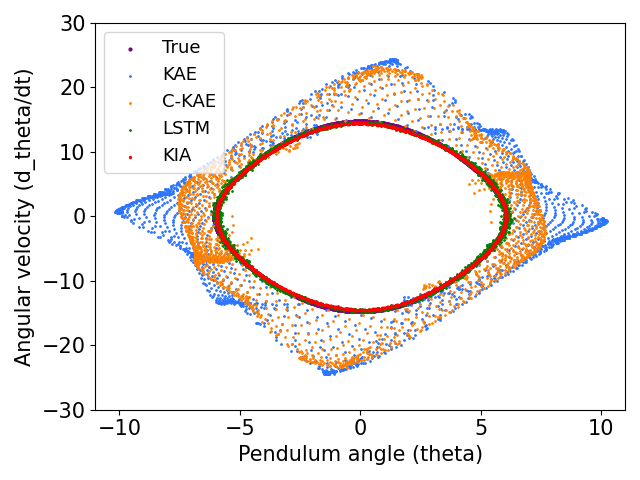}
        \label{fig:adapt}    
    \end{subfigure}
    \vspace{-0.2cm}
    \caption{\small Visualization of Prediction Trajectories on clean (top) and noisy input(bottom).}
    \label{fig:pendulum_trajectory}
    \vspace{-0.2cm}
\end{figure}

\paragraph*{Robustness comparison} Temporal physical processes often have noisy measurements due to sensor limitations and environmental conditions. We simulated the behavior by applying additive white gaussian noise with zero mean and varying standard. Specifically, to test model robustness, we evaluated KIA with perturbed inputs using small (0.1 std) and large (0.2 std) noise levels, and the results are depicted in Table \ref{tab:pendulum_results_2} and Table \ref{tab:pendulum_results_3} respectively. The noise disrupts the regular oscillation of the pendulum, causing slight deviations in its motion. Consequently, the system becomes more chaotic and difficult to accurately predict, resulting in increased errors as the noise levels rise. Interestingly, the KAE and C-KAE models shows more performance deterioration compared to the RNN and KIA models, especially over longer time horizons. Also, in table \ref{tab:pendulum_results_3}, we observe signs of overfitting in the KAE and C-KAE models, as indicated by lower errors in the initial $100$ predictions and higher errors in the last $100$ predictions. Figure \ref{fig:pendulum_prediction} illustrates the prediction error of the pendulum over 2,000 steps for clean and noisy observations. Comparing the RNN and KIA models, we find that KIA outperforms RNN, particularly over longer time horizons, even in the presence of noise. This highlights the robustness of KIA in effectively handling variations and disturbances in the input signals while capturing the true dynamics. 
\paragraph*{Effect of the size of the Training Data} Furthermore, to evaluate the influence of the labeled data size, we conducted experiments by utilizing different proportions of the training data on the top-performing RNN and KIA models. The results are presented in Table \ref{tab:varying_training_points}, where we compare the average prediction error obtained using subsamples of 200 and 300 instances from the training datasets. The KIA model consistently outperformed the RNN model in test error accuracy, even with fewer labeled data points, indicating its superior robustness and efficiency in using available data for precise predictions.

\paragraph*{Visualization of Prediction Trajectories} To gain further insights, we visualize the prediction trajectories of our models for an initial condition of $\theta_{0} = 2.4$. Figure \ref{fig:pendulum_trajectory} shows the trajectories generated using clean and noisy inputs, with the true trajectory shown in violet. Both KIA and RNN models are shown to  excel in capturing the pendulum dynamics, regardless of the presence of noise. The red and green lines align closely with the true trajectory, indicating their effectiveness. However, the predicted trajectories of the KAE and C-KAE models exhibit increasing deviations from the true trajectory, especially over longer time periods. The introduction of noisy inputs further amplifies these deviations, leading to a more pronounced divergence from the actual trajectory. This highlights the advantage of using KIA which utilize both forward and backward dynamics to learn underlying dynamics.

\subsection{Sea Surface Temperature Data}
\begin{figure}[!tbp]
    \centering
    \begin{subfigure}{.51\linewidth}
        \centering
        \includegraphics[width=\linewidth]{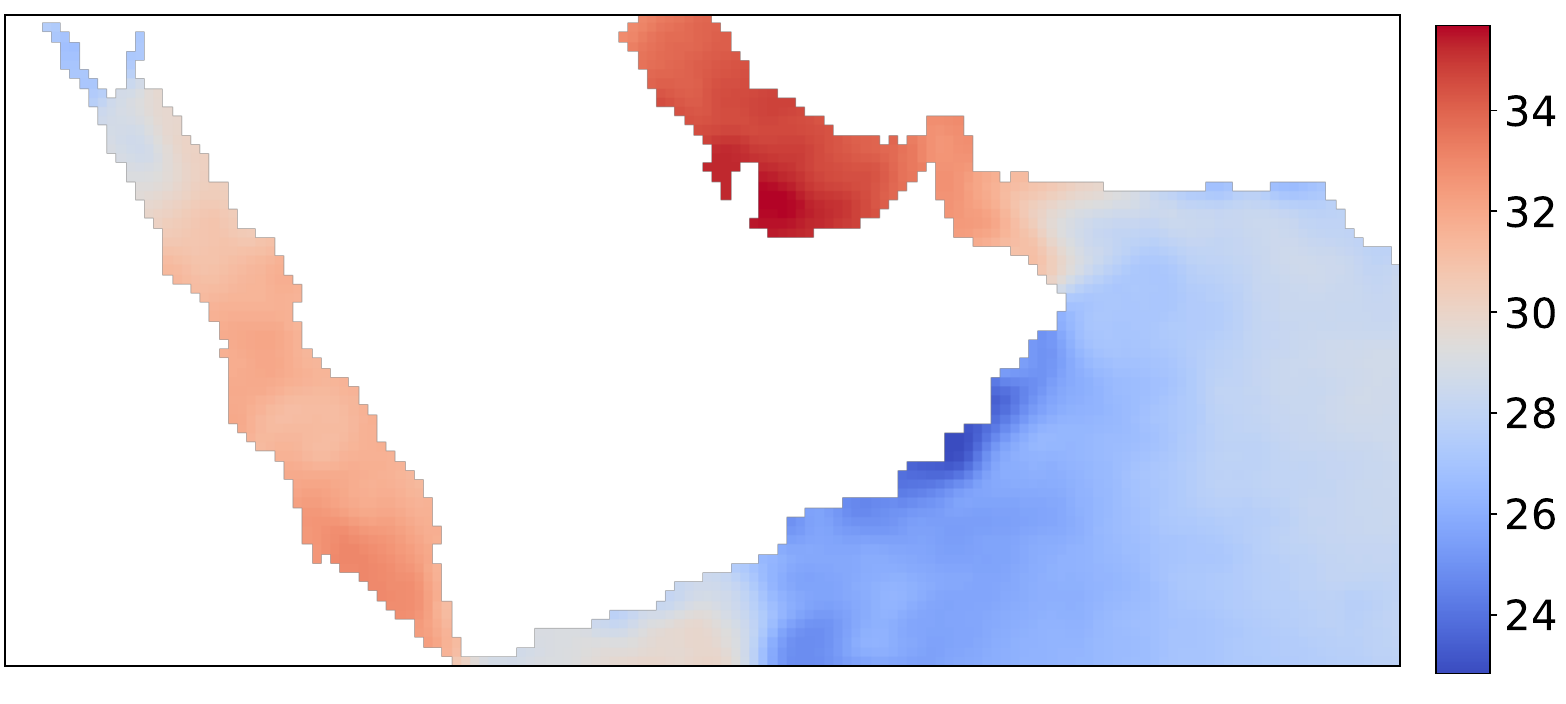}
    \end{subfigure}%
    \begin{subfigure}{.51  
    \linewidth}
        \centering
        \includegraphics[width=\linewidth]{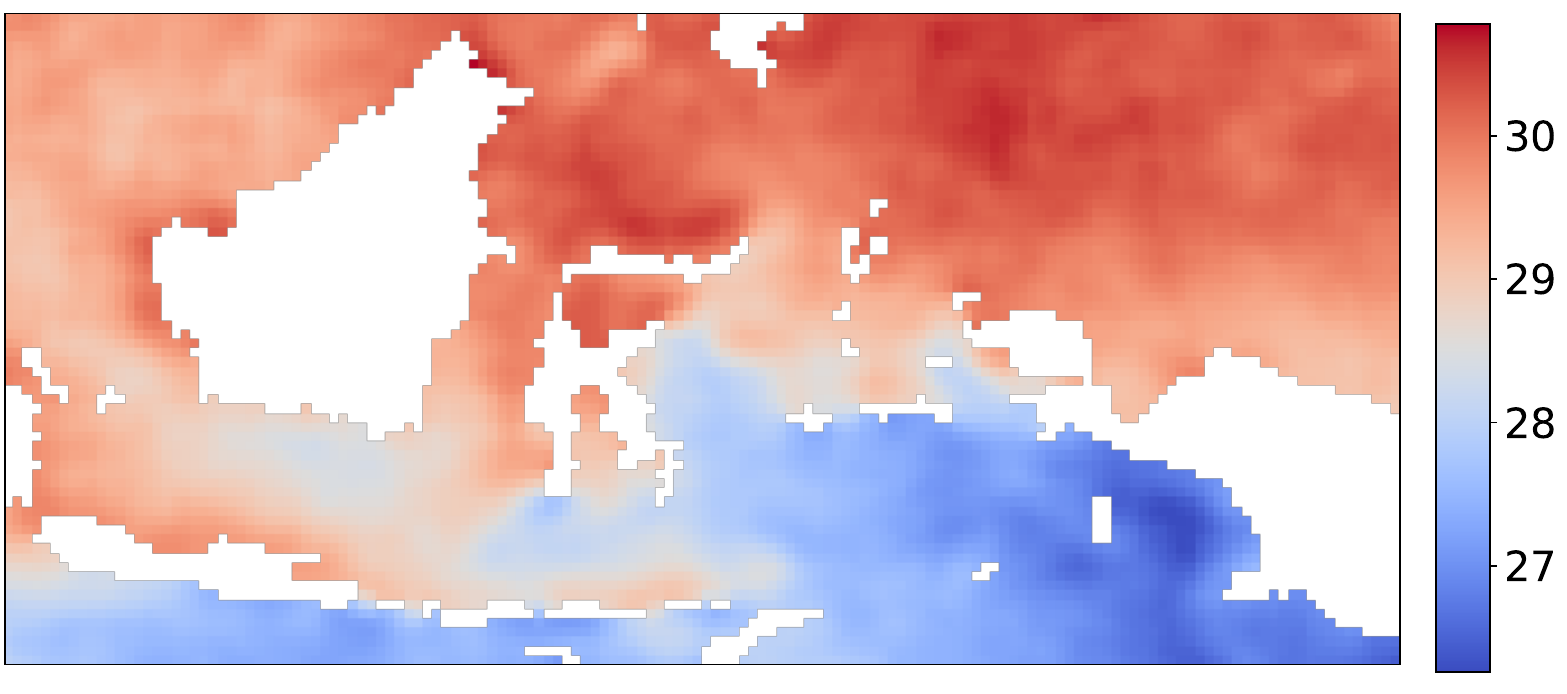}
       \end{subfigure}
    \caption{\small Example SST regions of the Persian Gulf (left) and Southeast Asia (right). The Persian Gulf, has a high temperature range of 23-35°C, displaying high variability, while Southeast Asia, has a range of 26-31°C, represents a low variability region. }
    \label{fig:SSTSnapshots}
\end{figure}
\textbf{Dataset Description:} In this experiment, we analyzed the NOAA Optimal Interpolation SST High Resolution dataset \cite{hurrell2008new}, which provides daily sea-surface temperature measurements at a spatial resolution of 0.25$^\circ$. %
We focused on two specific regions: the Persian Gulf (Lat: $12.5^\circ$ to $30^\circ$ N, Long: $31.25^\circ$ to $68.75^\circ$ E) and Southeast Asia (Lat: $-10^\circ$ to $7.5^\circ$ N, Long: $105^\circ$ to $142.5^\circ$ E) (see Fig \ref{fig:SSTSnapshots}). The Persian Gulf has higher sea surface temperatures and greater seasonal variations compared to Southeast Asia, making it more nonlinear from a modeling perspective. We used five years of data, with three years for training, one year for validation, and one year for testing. The performance of our models was evaluated using Celsius MAE as the error metric. This metric calculates the average absolute difference, in degrees Celsius, between the predicted and the actual temperatures, regardless of their direction (positive or negative). Forecasting climate patterns is an inherently difficult task \cite{chantry2021opportunities} due to the intricate interactions of a complex system. However, the input dynamics exhibit non-stationary periodic structures \cite{azencot2020forecasting}, suggesting the use of Koopman-based methods.

\begin{figure}[!tbp]
    \centering
    \begin{subfigure}{.51\linewidth}
        \centering
        \includegraphics[width=\linewidth]{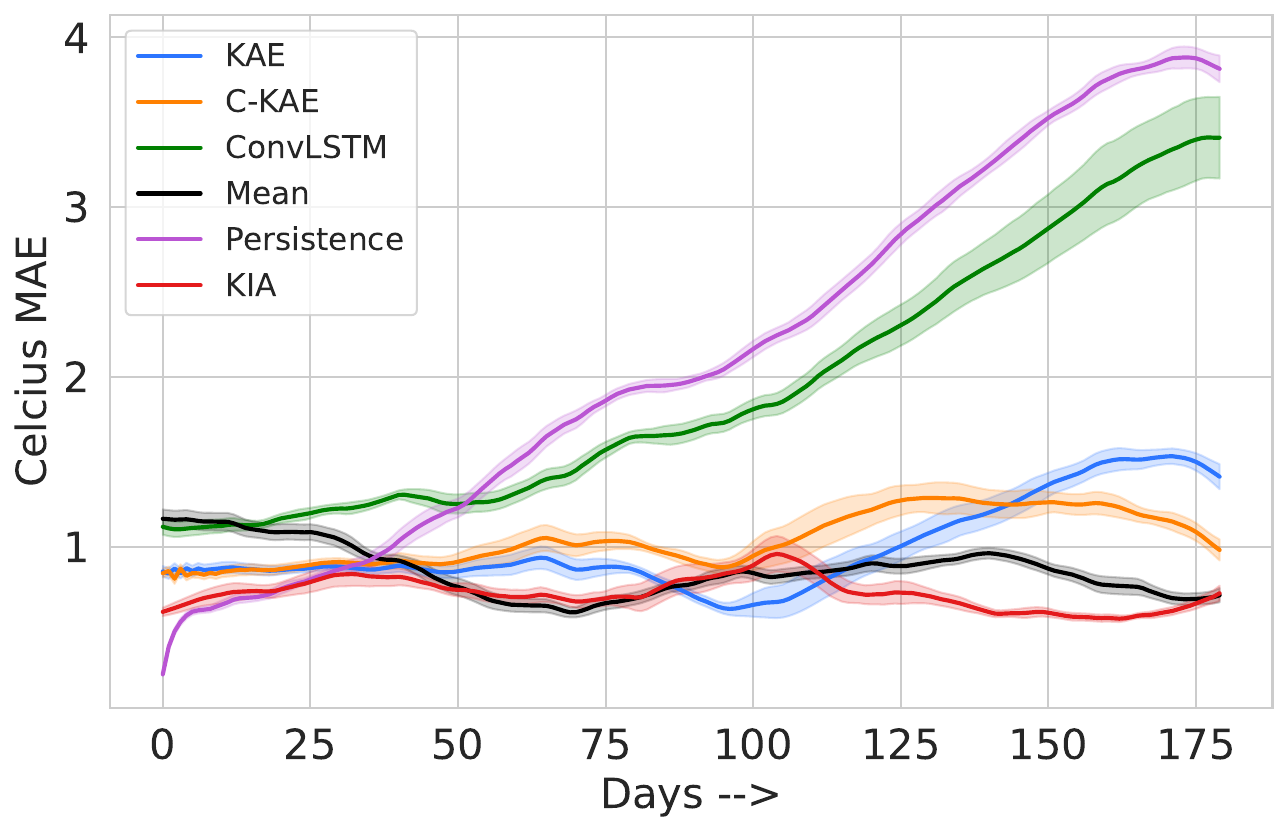}
    \end{subfigure}%
    \begin{subfigure}{.51  
    \linewidth}
        \centering
        \includegraphics[width=\linewidth]{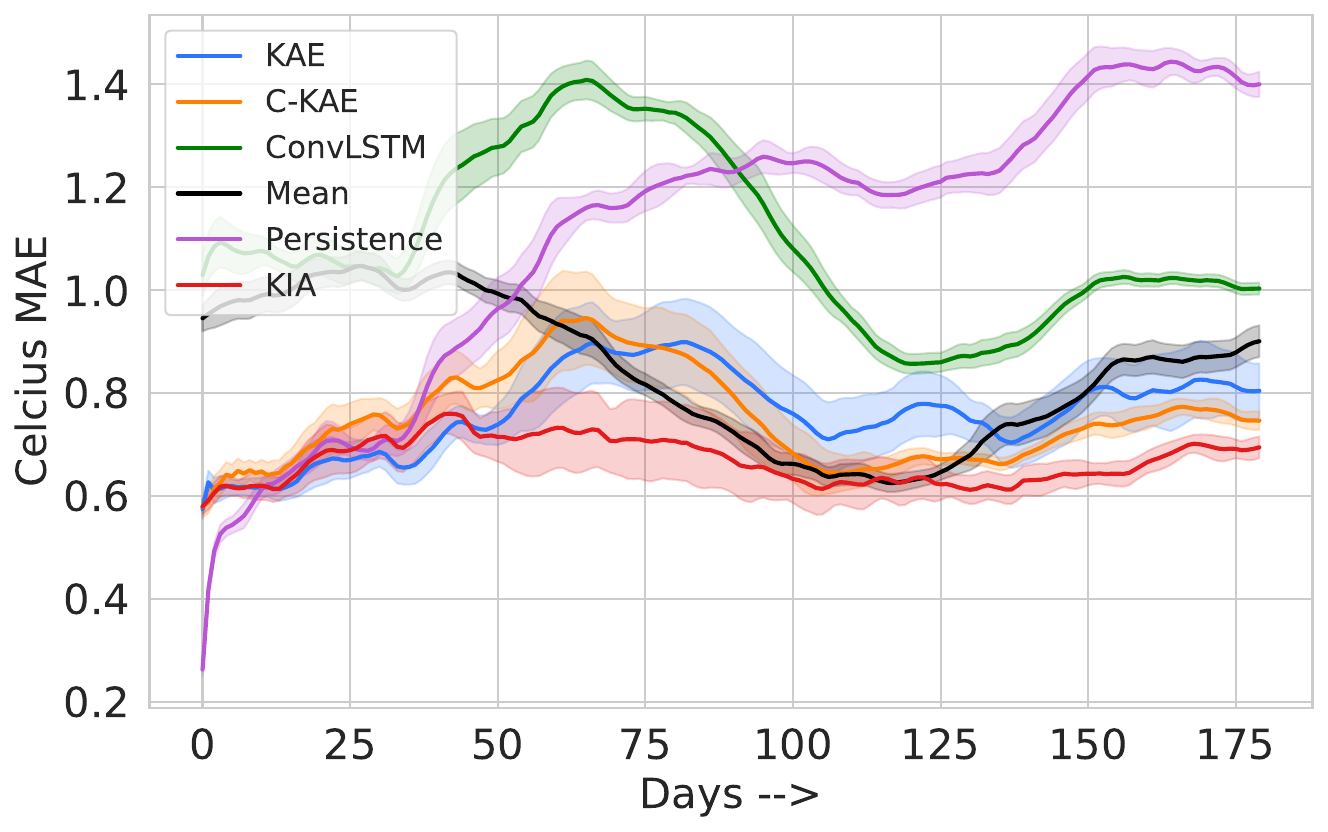}
       \end{subfigure}
    \caption{\small Prediction errors over a time horizon of 180 days for Persian Gulf (left) and Southeast Asia (right) }
    \label{fig:prediction_error_180days}
    \vspace{0.2cm}
\end{figure}

\begin{table}[]
\centering
\caption{\small  Sea surface temperature prediction error statistics: mean and standard deviation.}
\resizebox{0.38\textwidth}{!}{
\begin{tabular}{ccc}
\toprule
\textbf{Model} & \textbf{Regions} & \textbf{Prediction Error (Avg)} \\
\midrule
KAE & PG & 0.99 $\pm$ 0.06\\
C-KAE & PG & 1.04 $\pm$ 0.06\\
ConvLSTM & PG & 1.93 $\pm$ 0.10\\
Mean & PG & 0.86 $\pm$ 0.04\\
Persistence & PG & 2.13 $\pm$ 0.05\\
KIA & PG & \textbf{0.73} $\pm$ 0.05\\
\midrule
KAE & SE & 0.76 $\pm$ 0.06\\
C-KAE & SE & 0.75 $\pm$ 0.04\\
ConvLSTM & SE & 1.08 $\pm$ 0.03\\
Mean & SE & 0.84 $\pm$ 0.03\\
Persistence & SE & 1.10 $\pm$ 0.02\\
KIA & SE & \textbf{0.67} $\pm$ 0.04\\
\bottomrule 
\end{tabular}}
\label{tab:long_term_sst}
\end{table}

\begin{table}[]
\centering
\caption{\small Mean error of K-day ahead sea surface temperature prediction.}
\resizebox{0.5\textwidth}{!}{%
\begin{tabular}{lS[table-format=1.1]S[table-format=1.2]S[table-format=1.2]S[table-format=1.2]S[table-format=1.2]S[table-format=1.2]}
\toprule
\textbf{Model} & \textbf{Regions} & \multicolumn{5}{c}{\textbf{Prediction Error (Avg)}} \\
\cmidrule(lr){3-7}
& & {1 day} & {7 day}  & {14 day} & {21 day} & {30 day} \\  
\midrule
KAE & PG & 0.59 & 0.61 & 0.65 & 0.69 & 0.77 \\
C-KAE & PG & 0.56 & 0.60 & \textbf{0.63} & \textbf{0.67} & 0.76 \\
ConvLSTM & PG & 0.51 & 0.67 & 0.82 & 0.91 & 0.98 \\
Mean & PG & 0.81 & 0.81 & 0.81 & 0.81 & 0.81 \\
Persistane & PG & \textbf{0.21} & 0.59 & 0.78 & 0.96 & 1.19 \\
KIA & PG & 0.48 & \textbf{0.57} & \textbf{0.63} & \textbf{0.67} & \textbf{0.74} \\
\midrule
KAE & {SE} & 0.55 & 0.57 & 0.59 & 0.63 & 0.65 \\
C-KAE & {SE} & 0.54 & \textbf{0.56} & 0.59 & 0.62 & 0.64 \\
ConvLSTM & {SE} & 0.55 & 0.65 & 0.76 & 0.88 & 0.92 \\
Mean & {SE} & 0.75 & 0.75 & 0.75 & 0.75 & 0.75 \\
Persistance & {SE} & \textbf{0.24} & 0.61 & 0.68 & 0.72 & 0.77 \\
KIA & {SE} & 0.54 & \textbf{0.56} & \textbf{0.58} & \textbf{0.61} & \textbf{0.63} \\
\bottomrule 
\end{tabular}}
\label{tab:k_day_ahead_sst}
\end{table}

\begin{figure*}[h!]
    \centering
    \begin{subfigure}{0.38\linewidth}
        \centering
        \includegraphics[width=\linewidth]{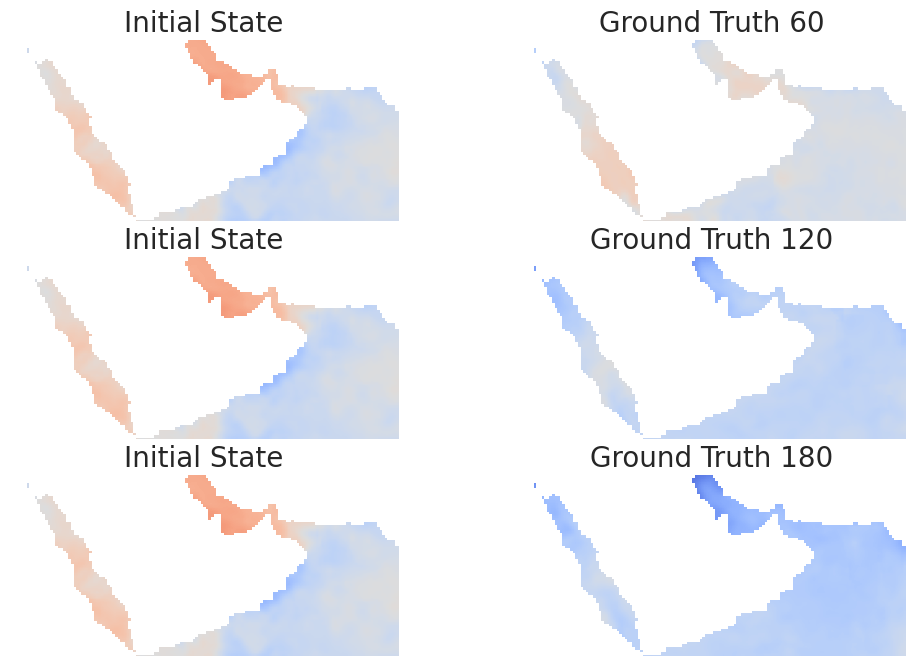}
    \end{subfigure}%
    \hspace{0.03\textwidth}
    \begin{subfigure}{0.57\linewidth}
        \centering
        \includegraphics[width=\linewidth]{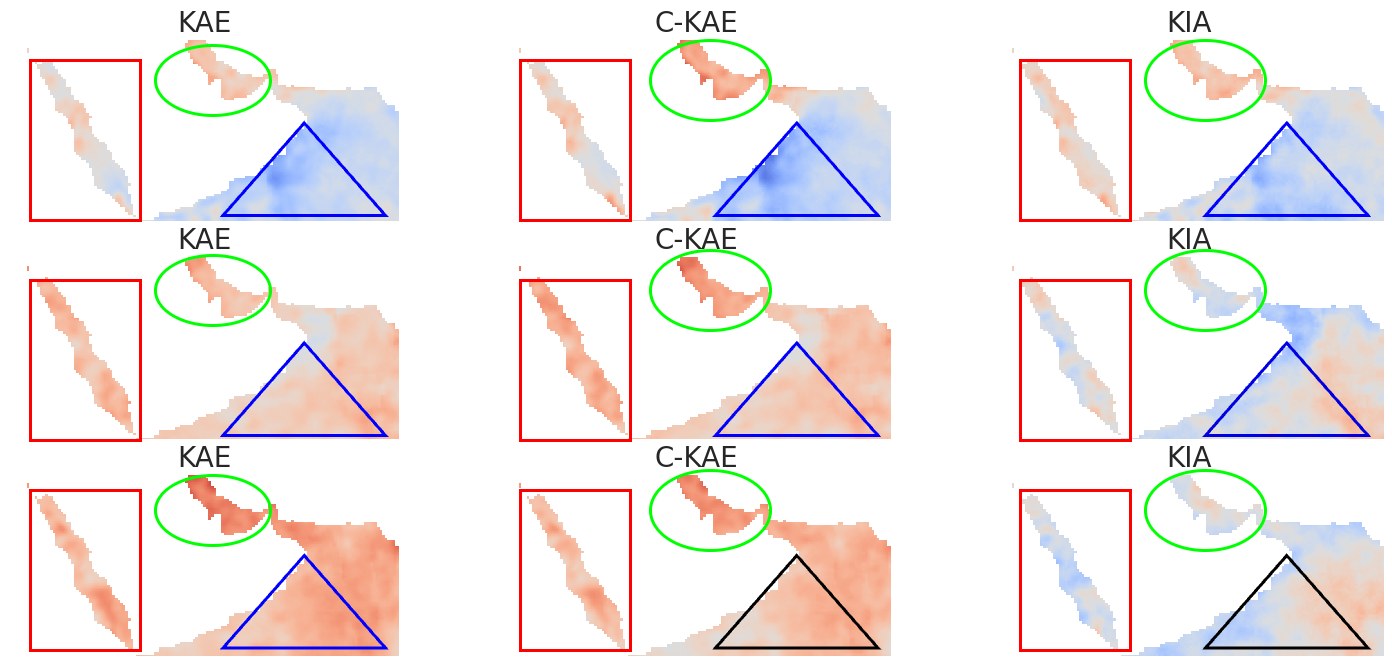}
    \end{subfigure}
    \caption{ \small We depict the future predictions for a single initial state over a horizon of 60, 120, and 180 days. The first column represents the consistent initial state, the second column shows the ground truth, and the remaining columns provide the model's predictions for days 60, 120, and 180. (best seen in color).}
    \label{fig:persiangulf_panel}
\end{figure*}

\begin{figure*}[h!]
    \centering
    \includegraphics[width=\linewidth]{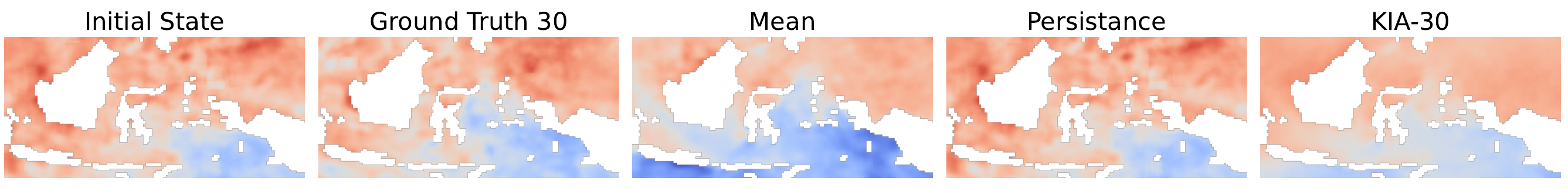}
    \caption{ \small We illustrate 30-day future forecasts for a random initial states. The first column shows the initial state, the second column presents the 30-day ahead ground truth, and the rest represent mean output, Persistence output, and KIA output.(best seen in color).}
    \label{fig:southeastasia_panel}
    \vspace{-0.5cm}
\end{figure*}

\textbf{Baselines:} We incorporates all Koopman-based baselines from Pendulum experiments, along with Persistence, Climate Mean, and replace LSTM with ConvLSTM. Persistence and Climate Mean are commonly used as baseline methods for climate predictions, offering different advantages based on the context and time scale. Persistence is effective for short-term forecasting, assuming stable climate conditions, while Climate Mean incorporates historical data for more reliable long-term predictions. Any useful model is expected to outperform these baselines. \underline{\textit{Persistence:}} This forecasting model assumes that future climate conditions will be similar to the current one, making identical predictions for each time step. \underline{\textit{Climate Mean:}} This model derives the average climate state from historical data of the same date. It suggests that the future climatic conditions will closely resemble the mean state derived from historical observations spanning previous years. \underline{\textit{ConvLSTM:}} is a spatiotemporal model specifically designed for data like SST, combining spatial processing with temporal dependencies using convolutional neural networks and LSTMs. It is widely adopted and used as an alternative to LSTM for spatio-temporal modeling.

\textbf{Experimental results: } In \textit{Long-term Prediction Testing} (as done in \cite{azencot2019consistent, rice2020analyzing}), we generate predictions for the next 180 days on 30 different initial days of the year to evaluate the accuracy and reliability of the model over an extended period. Table \ref{tab:long_term_sst} shows prediction error averages, and Fig.~\ref{fig:persiangulf_panel} visually compares different model outputs. The results show that our model's predictions for the 60-day, 120-day, and 180-day forecasts are closer to the ground truth compared to other models.  To enhance differentiation in the model's output, we use a color scheme that represents the temperature difference with the ground truth. Grey indicates a perfect prediction, while red and blue represent warmer and colder predictions, respectively. The initial state and ground truth temperatures are in Celsius. The oval, rectangle, and triangle regions were used for clear differentiation .The visual comparison shows that our model, KIA, performs better than CKAE and KAE models, which tend to overpredict. The prediction results show that KIA can generalize to unseen climate scenarios, which is significant considering the challenges involved in predicting climate data. Furthermore, Fig ~\ref{fig:prediction_error_180days} provides a visualization of the prediction error spanning a 180-day horizon, revealing interesting insights about different models' performance. While the persistence model is more accurate in short-term predictions, the KIA model outperforms the persistence and mean models when evaluating their overall performance. This indicates that the KIA model offers better long-term prediction capabilities, making it a more reliable choice for forecasting.

Additionally, we visualize the prediction error over the 180-day horizon in Fig ~\ref{fig:prediction_error_180days}, showing that while short-term predictions are more accurate with the persistence model, the KIA model outperforms both the persistence model and the mean model when considering overall performance.

\begin{figure}[!tbp]
    \centering
    \begin{subfigure}{.51\linewidth}
        \centering
        \includegraphics[width=\linewidth]{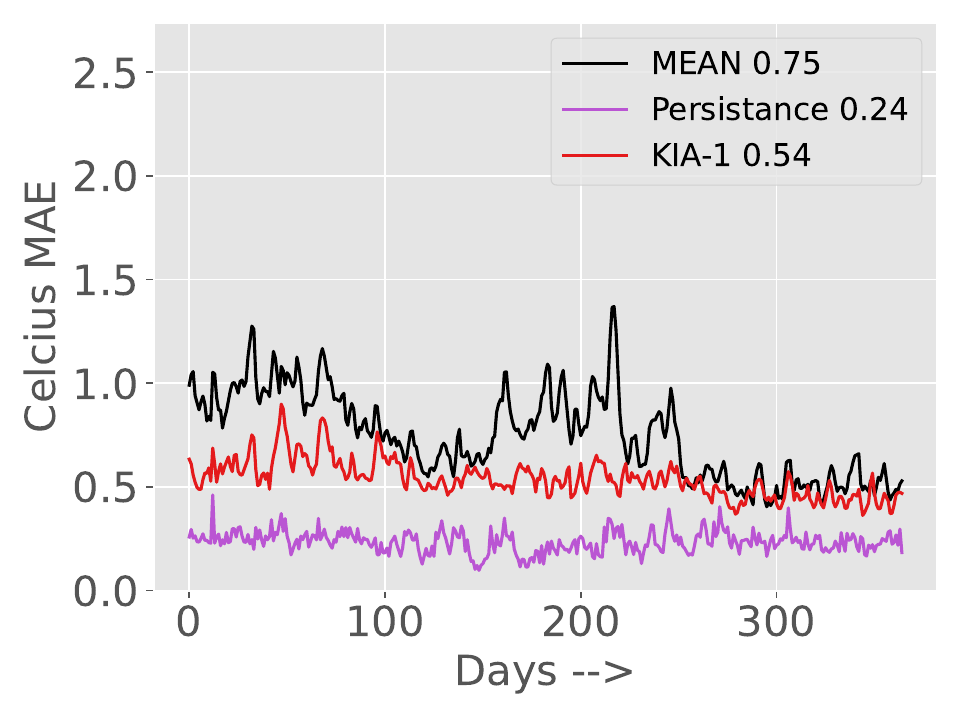}
    \end{subfigure}%
    \begin{subfigure}{.51  
    \linewidth}
        \centering
        \includegraphics[width=\linewidth]{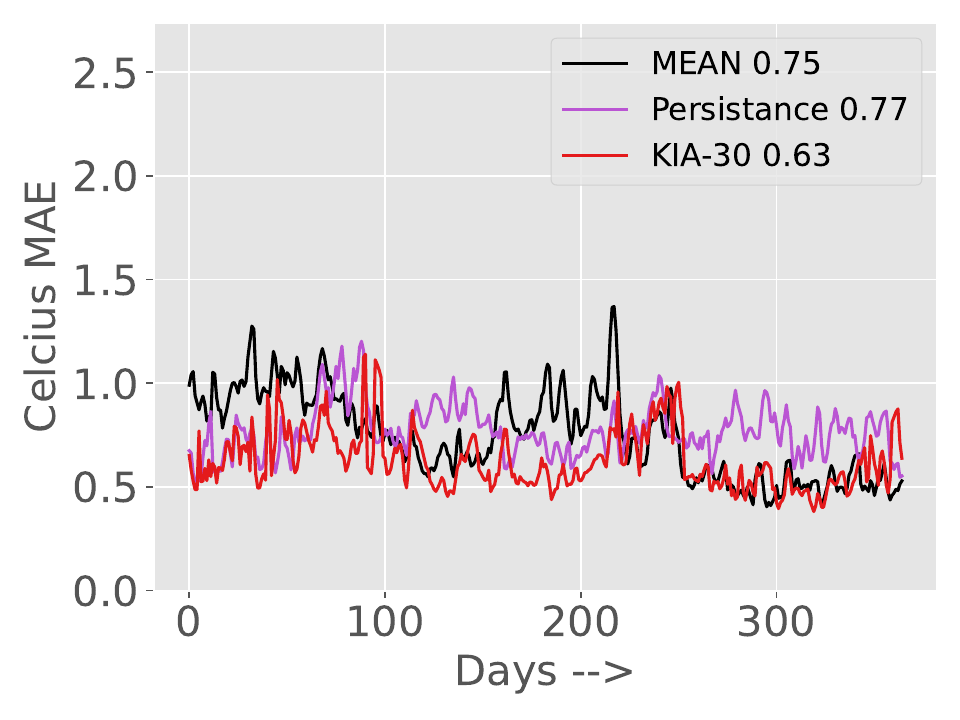}
       \end{subfigure}
    \caption{\small Prediction error for 1-day ahead (left) versus 30-day ahead (right) forecasts for southeast asia}
    \label{fig:predictio_error_year_sst}
    \vspace{0.2cm}
\end{figure}

\paragraph*{Impact of Timescales on K-Day Ahead Predictions}
In some days of a year, accurate forecasting can be more challenging due to factors like seasonality, irregular events, or volatility. Analyzing different timescales \cite{he2021sub} helps identify these days and assess their impact on model prediction. This information aids in resource allocation and the utilization of alternative forecasting methods during those periods. For this, we employ $K$ day ahead forecasts for each day in a  test year, where each prediction is based on input from $K$ days in the past.  In our case, we evaluated our model for $K$ in \{1, 7, 14, 21 and 30\} days. Table \ref{tab:k_day_ahead_sst} showcases the mean prediction error values (standard deviation is omitted for brevity). We observe that all Koopman-based models do an equally good job for short-term prediction. Also, the persistence model is the best if we are interested in a 1-day ahead forecast. However, even in the short term, from K=7 days onwards, our approach surpasses the persistence and mean models. Fig.~\ref{fig:southeastasia_panel} displays the 30-day KIA forecast for randomly selected initial states, which is visually compared with the mean and persistence models. We can observe that for  K=30, the mean is overpredicting, persistence is underpredicting, and KIA output is considerably closer to the actual data. The prediction results are highly significant, as successfully predicting climate data and surpassing the mean and persistence models in the short term is challenging. Fig.~\ref{fig:predictio_error_year_sst} further visualizes the prediction error for the entire year for $K = 1$ and $K = 30$ day. From the $K=30$ plot, we can identify that the later part of the years is difficult, where the models struggle to provide better predictions than the mean. In addition, from the plot, we can also identify dates like the $150^{th}$ day of the year, where the mean is doing better than all the models. From these results, we can comprehend that the KIA can capture the true essence of the system dynamics and can be helpful for domain scientists for long-term prediction. The code used in this paper is available at Google Drive 
\footnote{\tiny \url{https://drive.google.com/drive/folders/1_vwNigBBzRZRpduFluu1mgBNXTcm1htx}}.

\section{Discussion and Conclusion}

This paper proposes KIA, a long-term predictive model for temporal data. Our approach is built upon the principles of the Koopman theory, which enables the approximation of dynamical systems using linear evolution matrices. The key contribution of our method lies in its incorporation of both forward and backward dynamics, achieved through a meticulously designed invertible neural network architecture that ensures consistency between these two directions. We successfully demonstrated the model's capability to accurately capture long-term trends, recover underlying dynamics, and generalize to unseen scenarios through experiments on pendulum and climate datasets. Although our improvements have been validated using these specific datasets, our architecture's versatility allows for its application in various other forecasting tasks that involve backward dynamics. For instance, consider the video captioning task \cite{zhang2019object}, where the temporal order of video frames in both forward and backward directions plays a crucial role in capturing the movements of significant objects from different perspectives. Previous studies have employed graph-based networks to capture the relationships between objects in videos. In this context, our approach presents an alternative solution, thereby enabling a comprehensive understanding and captioning of the visual content. Furthermore, our method exhibits potential \cite{liu2020learning} for integration within Transformer models \cite{wolf2020transformers}, a popular architecture widely used for sequence modeling tasks. The integration holds promise for addressing more complex temporal problems across diverse domains, further improving the accuracy and generalizability of forecasting systems.

\bibliographystyle{plain}

\end{document}